\documentclass{article}

\usepackage{arxiv}

\usepackage[utf8]{inputenc} 
\usepackage[T1]{fontenc}    
\usepackage{hyperref}       
\hypersetup{
    colorlinks,
    citecolor=blue,
    filecolor=black,
    linkcolor=blue,
    urlcolor=black
}
\usepackage{url}            
\usepackage{booktabs}       
\usepackage{amsmath,amssymb,amsfonts}       
\usepackage{nicefrac}       
\usepackage{microtype}      
\usepackage{cleveref}       
\usepackage{lipsum}         
\usepackage{graphicx}
\usepackage{doi}

\usepackage{IEEEtrantools}
\usepackage{soul}
\usepackage{cite}
\usepackage{float}
\usepackage{caption}
\usepackage{subcaption}
\usepackage{stfloats}
\usepackage{enumitem}
\usepackage{mathtools}
\usepackage{algorithm}
\usepackage{algpseudocode}
\usepackage{graphicx,color}
\usepackage{textcomp}
\usepackage{xcolor}
\usepackage{bm}
\usepackage{diagbox}
\usepackage{multirow}
\usepackage{makecell}
\usepackage{cases}
\usepackage{tikz}
\usepackage[british]{babel}
\usepackage[mathscr]{eucal}
\newcommand\norm[1]{\left\lVert#1\right\rVert}

\title{Physics Informed Deep Learning: Applications in Transportation}

\date{}

\author{
    \href{https://orcid.org/0000-0001-6736-5627}{\includegraphics[scale=0.06]{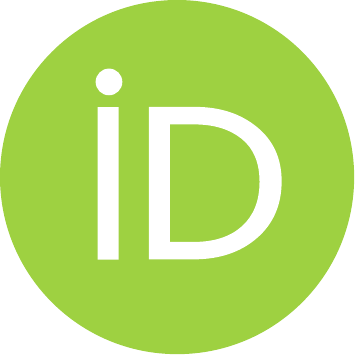}\hspace{1mm}Archie J.~Huang}\,,
    \href{https://orcid.org/0000-0001-7754-6341}{\includegraphics[scale=0.06]{figures/orcid.pdf}\hspace{1mm}Shaurya~Agarwal}\\
    UrbanITY Lab\\
    Civil, Environmental \& Construction Engineering Department\\
    University of Central Florida\\
    Orlando, FL 32816, USA \\
    \texttt{archie.huang@nyu.edu, shaurya.agarwal@ucf.edu} \\
}

\renewcommand{\shorttitle}{PIDL Applications in Transportation}

\hypersetup{
pdftitle={Physics Informed Deep Learning: Applications in Transportation},
pdfsubject={PIDL Applications in Transportation},
pdfauthor={Archie J.~Huang, Shaurya~Agarwal},
pdfkeywords={Traffic State Estimation, Physics-informed Deep Learning, Traffic Flow Theory, Conservation Law},
}

\begin{document}
\maketitle

\begin{abstract}
A recent development in machine learning - physics-informed deep learning (PIDL) - presents unique advantages in transportation applications such as traffic state estimation. Consolidating the benefits of deep learning (DL) and the governing physical equations, it shows the potential to complement traditional sensing methods in obtaining traffic states. In this paper, we first explain the conservation law from the traffic flow theory as ``physics'', then present the architecture of a PIDL neural network and demonstrate its effectiveness in learning traffic conditions of unobserved areas. In addition, we also exhibit the data collection scenario using fog computing infrastructure. A case study on estimating the vehicle velocity is presented and the result shows that PIDL surpasses the performance of a regular DL neural network with the same learning architecture, in terms of convergence time and reconstruction accuracy. The encouraging results showcase the broad potential of PIDL for real-time applications in transportation with a low amount of training data.
\end{abstract}

\keywords{Traffic State Estimation \and Physics-informed Deep Learning \and Traffic Flow Theory \and Conservation law}

\section{Introduction and Motivation}\label{sec:int}

Development in deep learning (DL) neural networks \cite{shi2021physics2} \cite{shi2021physics} benefits a wide range of engineering applications. The learning capability of a DL neural network helps practitioners in numerous fields such as transportation engineering and has been widely adopted in projects on object detection, autonomous driving, and estimations of system conditions.    

Traffic state estimation (TSE) is a crucial task for transportation planners in understanding travel demand and road infrastructure's level of service (LOS). Due to the cost constraints associated with installing sensing devices along freeways and arterial roads, traffic observations can solely be obtained at predetermined locations, leaving areas where traffic conditions are unperceived. Traffic states such as vehicle speed $v$, density $\rho$, and flow $f$ in the unobserved regions need to be approximated by using the collected measurements of traffic at sparse locations \cite{seo2017traffic}. Take loop detectors as an example: the signal indicating the passage of a vehicle can only be obtained at predetermined locations where the electrically conducting loops are planted. The number of detectors deployed considerably affects the quantity of traffic data collected from a highway system \cite{herrera2010incorporation}. 

The task of TSE is further impeded by issues such as the measurement noise in detectors and data loss due to sensor malfunctions \cite{bekiaris2016highway} \cite{contreras2017quality} \cite{contreras2015observability}. The inaccuracy in recorded traffic data and the limited data resolution during signal processing contribute to the challenges in precise TSE \cite{agarwal2015dynamic} \cite{agarwal2019controllability}. 

\begin{figure}[htbp]
    \begin{center}
        \includegraphics[width=0.6\textwidth]{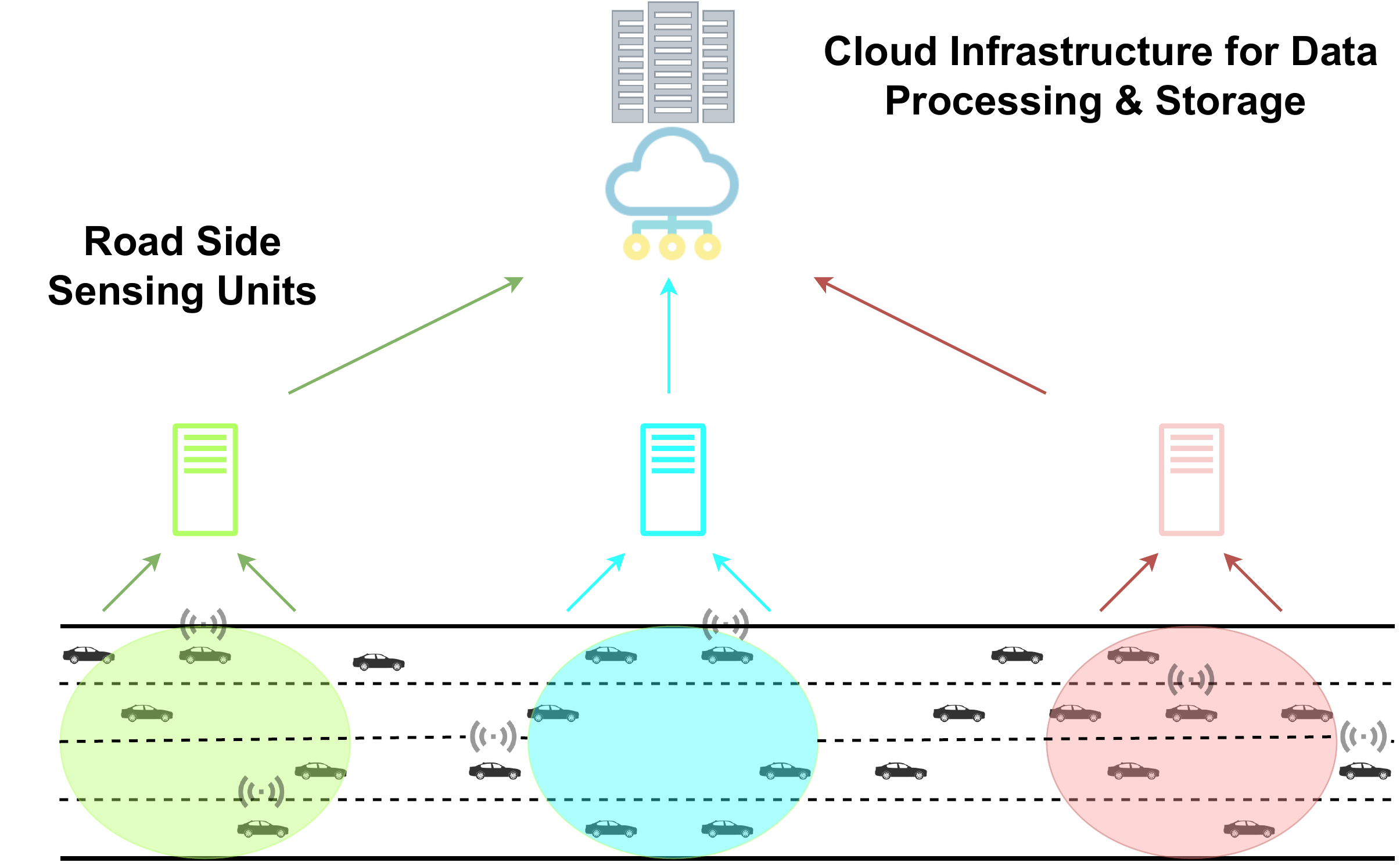}
        \caption{Process of Traffic State Data Acquisition}
        \label{fig:inf}
    \end{center}
\end{figure}

Fig.~\ref{fig:inf} illustrates the process of traffic state data acquisition: sensing devices collect information such as speed $v$ and headway $T$ at designated locations and broadcast the information to central cloud infrastructure for data processing and storage. Note that the location of sensors (loop detectors, radars, or roadside units (RSUs) only provides sparse measurement, and traditionally model-based methods such as Kalman filter have been used to estimate traffic conditions in regions without observed data. However, the model-driven approaches are restricted by the accuracy of the model and its chosen parameters.

Development in DL such as long short-term memory (LSTM) and recurrent neural networks (RNNs) have successfully captured the underlying relationships among traffic states \cite{rahman2018short}, showing promise in TSE. Nevertheless, the DL approaches are data-hungry and computationally intensive. Readers are redirected to \cite{huang2022physics} for a detailed discussion on state-of-the-art TSE approaches. Given the sparse observation of traffic, it may also be prone to the overfitting issue. 

\textbf{Research questions:} A careful analysis of the aforementioned literature prompts the following research question: can governing physical equations come to aid the data-hungry DL neural network, and improve its estimation accuracy by preempting overfitting? 

This paper presents details of physics-informed deep learning (PIDL), which integrates the cost terms of a regular neural network with the underlying relationship between state variables, and displays PIDL application in TSE using the Lighthill-Whitham-Richards (LWR) conservation law. The work is inspired by the development of training DL neural networks with the aid of governing differential equations \cite{raissi2017physics} \cite{raissi2019physics} and built upon our previous research on traffic state estimation \cite{huang2022physics} \cite{huang2020physics}. The results show great potential in TSE applications to obtain traffic conditions in unperceived areas.

Here is how the rest of the paper organized: Section~\ref{sec:ptf} examines traffic flow theory and furnishes knowledge on deep neural networks. Section~\ref{sec:pid} explains how to add physics into the training process of a DL neural network for TSE. Section~\ref{sec:dat} discusses the scenario of traffic data collection. Section~\ref{sec:cs} explores a case study with velocity data to test the performance of PIDL. Finally, section~\ref{sec:C} concludes the paper and proposes future directions.

\section{Physics of Traffic Flow and Deep Learning}\label{sec:ptf}
\subsection{Traffic Flow Fundamentals}

Traffic state variables describe the spatial-temporal knowledge of road conditions. Lighthill-Whitham-Richards (LWR) model \cite{richards1956shock} portrays the relationship between three traffic state variables - the cumulative count of vehicles $N(x,\; t)$, density $\rho(x,\; t)$ and flow $q(x,\; t)$. If $N(x, t)$ is differentiable with respect to both location $x$ and time $t$, the conservation law of traffic flow is given by \eqref{eqn:lwr}.

\begin{equation} \label{eqn:lwr}
 \frac{\partial q(x,\; t)}{\partial x} + \frac{\partial \rho(x,\; t)}{\partial t} = \frac{\partial^2 N(x, \; t)}{\partial t \partial x} - \frac{\partial^2 N(x, \; t)}{\partial x \partial t} = 0
\end{equation}

\begin{figure}[htbp]
   \begin{center}
        \includegraphics[width=0.6\textwidth]{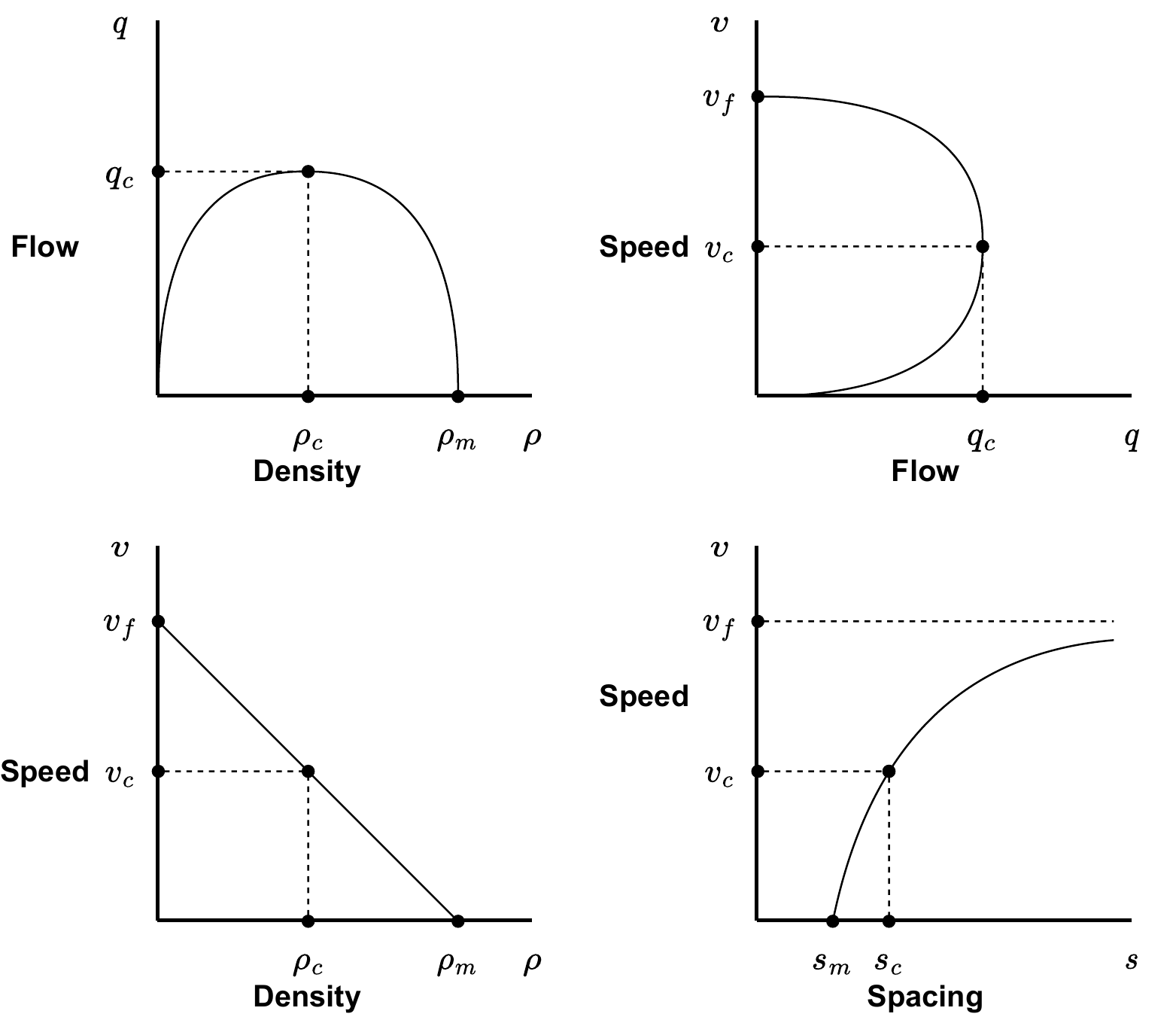}
        \caption{Greenshields' Fundamental Diagram}
        \label{fig:gre}
   \end{center}
\end{figure}

Greenshields' fundamental diagram (FD) depicts the relationship between speed $v$, density $\rho$, and flow $q$ \cite{greenshields1935study}. The connection between traffic state variables is formulated in \eqref{eqn:green}, in which $\rho_{max}$ is the maximum density and $v_{free}$ is the maximum speed (also known as the free-flow speed). Greenshields' FD is shown in Fig.~\ref{fig:gre}.

\begin{equation} \label{eqn:green}
\begin{split}
q(\rho) &= v_{free} \left(1 - \frac{\rho}{\rho_{max}}\right) \rho \\
v(\rho) &= v_{free} \left(1 - \frac{\rho}{\rho_{max}}\right) \\
\end{split}
\end{equation}

\subsection{Deep Learning Neural Network}

A deep learning (DL) neural network includes several layers of computation neurons. To measure the disagreement between the prediction or estimation produced by a DL neural network and the ground truth, a cost function is configured and minimizing the cost is the objective in training a DL neural network. Adopting mean square error (MSE) as the cost measure, we can build a cost function of the neural network, termed as the DL-cost $J_{DL}$. The DL-cost is written in \eqref{eqn:cost}, in which $\hat{v}(x,\; t)$ is the estimation of the velocity $v(x,\; t)$ at location $x$ and time $t$. $N_1$ is the number of spatial bins on the road segment, and $N_2$ is the number of temporal bins of time $t$.

\begin{equation} \label{eqn:cost}
    J_{DL} = MSE_{(\hat{v}(x, t), v(x, t))} = \frac{1}{N_1 \cdot N_2}\sum_{i=1}^{N_{1}}\sum_{j=1}^{N_{2}}\lvert\hat{v}(x^{(i)}, t^{(j)}) - v(x^{(i)}, t^{(j)})\rvert^2
\end{equation}

As the training process drives to minimize cost, the accuracy of the neural network is assessed based on the accuracy of the state estimation output. The Frobenius norm is selected to build a normalized error measurement $\mathit{E}$. It is shown as \eqref{eqn:acc}, in which $\mathbf{V}$ is the matrix form of velocity $v(x, \; t)$, and $\mathbf{\hat{V}}$ is the neural network output (estimation) of $\mathbf{V}$. $\mathit{E}$ normalizes the cost $J_{DL}$ and represents the relative estimation error. $N_1$ and $N_2$ are the numbers of discretized spatial and temporal bins of the velocity field.   

\begin{equation} \label{eqn:acc}
    E = \frac{\norm{\mathbf{V} - \mathbf{\hat{V}}}_F}{\norm{\mathbf{V}}_F} \times 100\%  = \frac{\sqrt{\sum_{i=1}^{N_{1}}\sum_{j=1}^{N_{2}}\lvert\hat{v}(x^{(i)}, t^{(j)}) - v(x^{(i)}, t^{(j)})\rvert^2}}{\sqrt{\sum_{i=1}^{N_{1}}\sum_{j=1}^{N_{2}}\lvert v(x^{(i)}, t^{(j)})\rvert^2}} \times 100\% 
\end{equation}

\section{Physics-informed Deep Learning}\label{sec:pid}

The LWR conservation law is a partial differential equation (PDE). Since fundamental diagrams relate the traffic state variables, we can use the equations of Greenshields' FD in \eqref{eqn:green} to transform LWR in \eqref{eqn:lwr} to \eqref{eqn:lwr2}, which contains only one independent variable, velocity $v(x, \; t)$.

\begin{equation} \label{eqn:lwr2}
	\rho_{max} \left(1 - \frac{2v(x, \; t)}{v_{free}}\right)  \frac{\partial v(x, \; t)}{\partial x}   - \frac{\rho_{max}}{v_{free}} \frac{\partial v(x, \; t)}{\partial t} = 0
\end{equation}

We establish two measures: (a) DL-cost, $J_{DL}$, for the estimation error between the output and the ground truth, and (b) physics-cost, $J_{PHY}$, for the noncompliance of conservation law. Adopting the LWR conservation law in \eqref{eqn:lwr2} as ``physics'', these two cost terms are formulated in \eqref{eqn:mea2}. The number of observed data is denoted as $N_{0}$, and $N_{c}$ is the number of the collocation points where the measurement of non-compliance of physics - $J_{PHY}$ is assessed.  

\begin{equation} \label{eqn:mea2}
	\begin{split}
	J_{DL} &= \frac{1}{N_{o}}\sum_{j=1}^{N_{o}}\lvert v(x_{o}^{j}, \; t_{o}^{j}) - \hat{v}(x_{o}^{j}, \;  t_{o}^{j})\rvert^2 \\
	J_{PHY} &= \frac{1}{N_{c}}\sum_{j=1}^{N_{c}}\lvert \rho_{max} (1 - \frac{2\hat{v}(x_{c}^{j}, \; t_{c}^{j})}{v_{free}})\frac{\partial \hat{v}(x_{c}^{j}, \; t_{c}^{j})}{\partial x} - \frac{\rho_{max}}{v_{free}} \frac{\partial \hat{v}(x_{c}^{j}, \;t_{c}^{j})}{\partial t}\rvert^2 
	\end{split}
\end{equation}

By adding the physics-cost $J_{PHY}$ to the cost function, the LWR conservation law in \eqref{eqn:lwr} is integrated into the training process of the neural network. Now the neural network is able to use the governing physical equations as the ``knowledge of traffic flow''. The physics aids in the process of finding the best-fitting parameters of the neural network and prevents overfitting, leads to better performance on TSE. The physics-cost term can be adjusted by introducing a weight hyperparameter $\alpha$. The entire cost function is given as \eqref{eqn:final}. 

\begin{equation} \label{eqn:final}
	\mathcal{J} = J_{DL} + \alpha * J_{PHY}
\end{equation}

\begin{figure}[htbp]
   \begin{center}
        \includegraphics[width=0.75\textwidth]{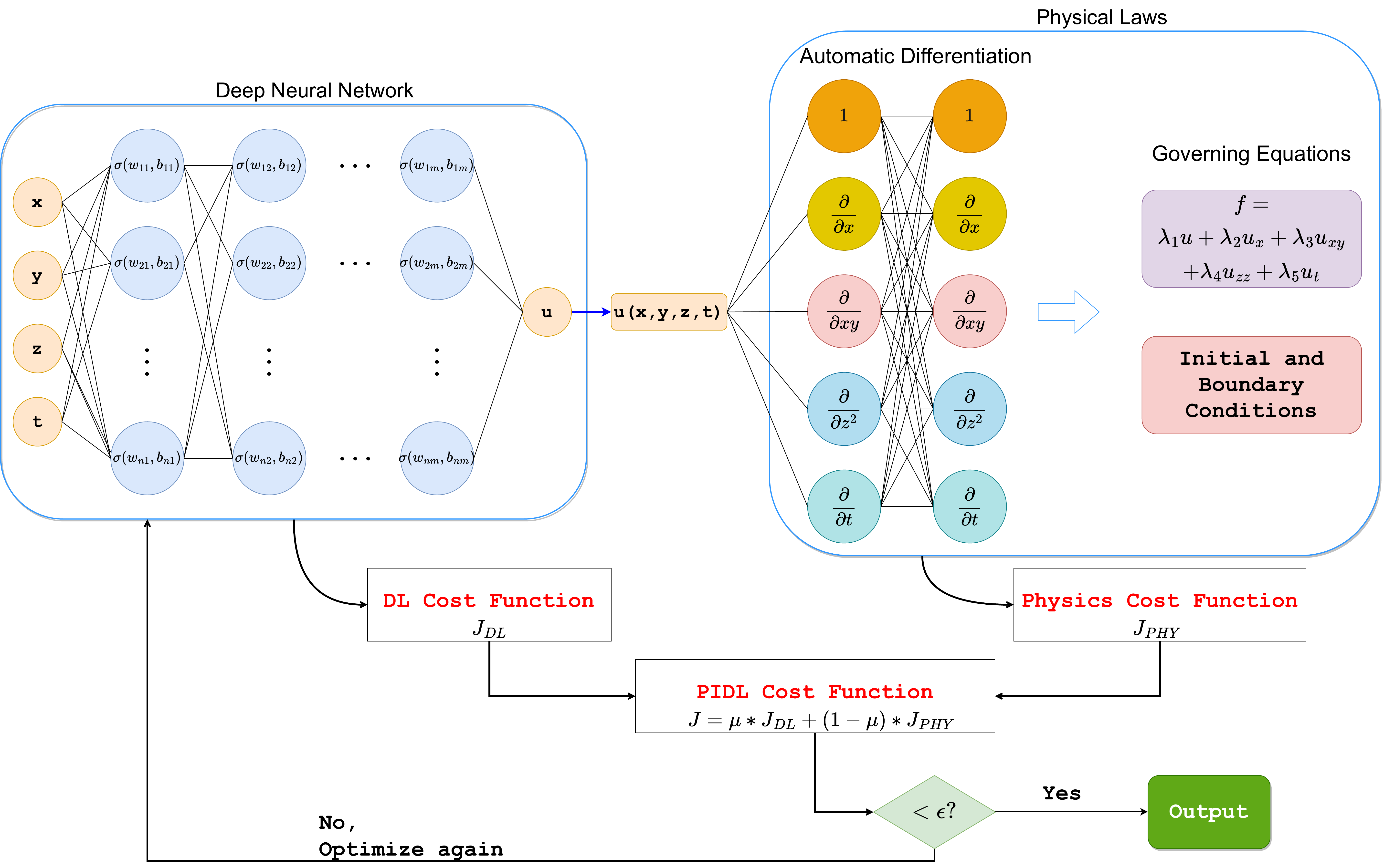}
        \caption{Architecture of a PIDL Neural Network}
        \label{fig:pinn}
   \end{center}
\end{figure}

The architecture of a PIDL neural network is shown in Fig.~\ref{fig:pinn}. It equips a regular DL neural network with a training process that involves minimizing the combined cost of $J_{DL}$ and $J_{PHY}$, given in \eqref{eqn:final}. By minimizing the combined cost $\mathcal{J}$, the neural network learns to calibrate its parameters based on the observed data and the physics. Once $\mathcal{J}$ is smaller than the predetermined threshold, it gives the fine-tuned output as the final estimation. 

\section{Data Collection Scenario} \label{sec:dat}

In this section, we paint the picture of traffic data collection scenario. On a transportation corridor, there is abundant processing power untapped or generally under-utilized in on-site edge nodes, also known as cloudlets \cite{satyanarayanan2009case} or fog nodes \cite{luan2015fog}. They are placed at the edge of the communication network and in close proximity to the end devices where traffic data originate \cite{satyanarayanan2017emergence}. Edge devices such as local traffic controllers, roadside units can receive and record traffic data reported by detectors along the road or smart onboard devices in connected vehicles (CV) regarding vehicle location and velocity. 

If deployed intermittently, edge devices such as RSUs may not locate within the communication range, restricting data sharing between the devices. Fog infrastructure provides a fitting solution to this obstacle by connecting the otherwise isolated RSUs. It harnesses the data collection capability of the RSUs and serves as an intermediate layer between edge devices and the cloud server.

\begin{figure}[htbp]
   \begin{center}
        \includegraphics[width=0.75\textwidth]{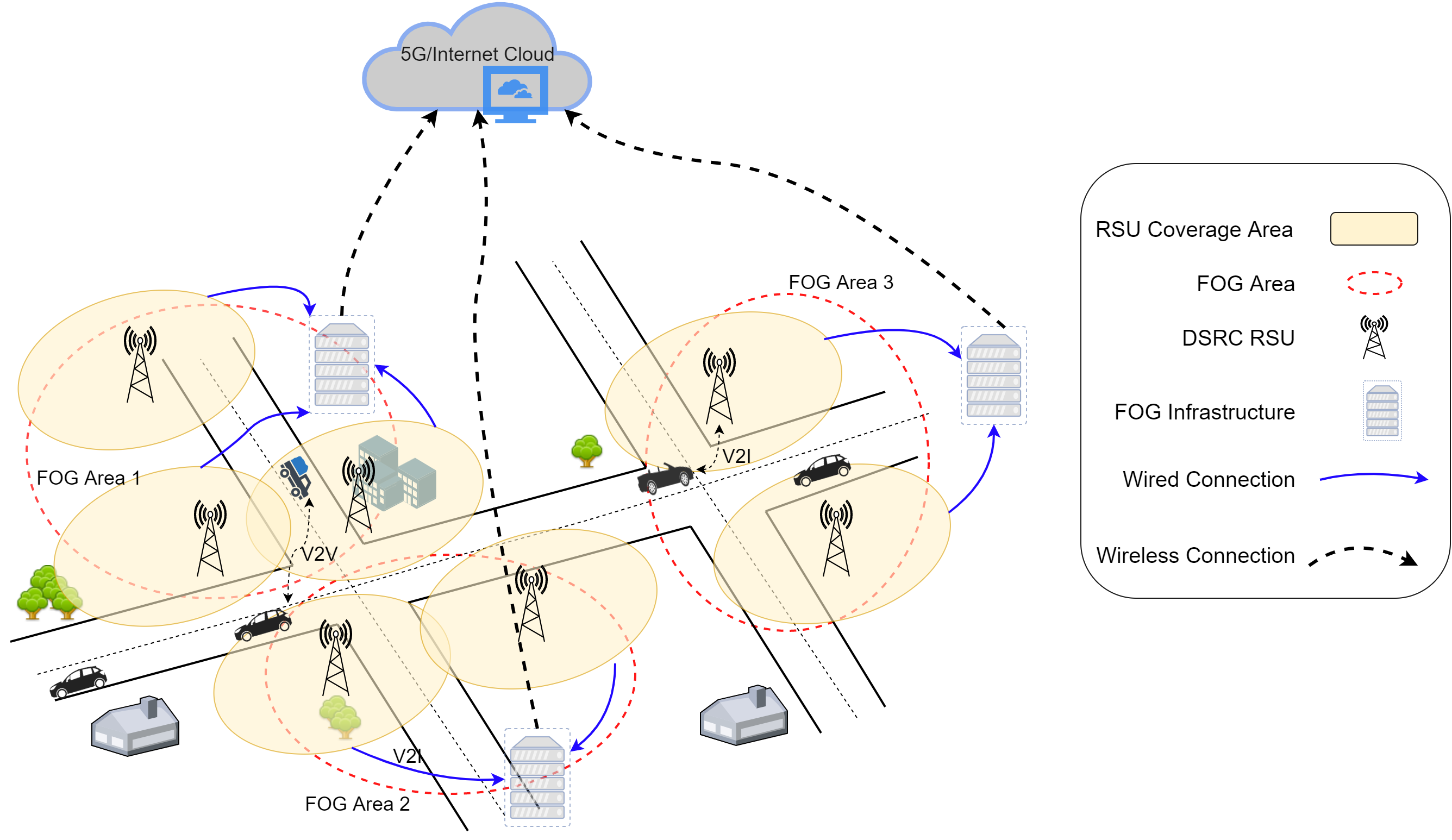}
        \caption{Data Collection Scenario using Fog Architecture}
        \label{fig:fog}
   \end{center}
\end{figure} 

In other words, edge computing pushes the brainpower and communication capabilities into the end device via programmable logic controllers and programmable automation controllers. The concept of fog computing was developed to harness the data collection, storage, and computing power of end devices and serve as an intermediate layer between edge layer and cloud servers --- thus decreasing network latency and improving system response time \cite{bonomi2012fog}. An application scenario employing fog infrastructure is illustrated in Fig.~\ref{fig:fog}.  

\section{Case Study}\label{sec:cs}

A DL neural network and a PIDL neural network with the same learning architecture are created for the case study. Using the Lax-Hopf method \cite{mazare2011analytical}, a synthetic traffic dataset is created under no upstream and downstream flow conditions. The value of velocity $v(x, \; t)$ is simulated at $x \in [0, 5000]$ meters and during $t \in [0, 240]$ seconds. The road segment is evenly divided into 500 spatial bins $(\delta x = 10m)$ and time $t$ is evenly divided into 240 units $(\delta t = 1s)$.

\begin{figure}[htbp]
    \begin{center}
        \includegraphics[width=0.5\textwidth]{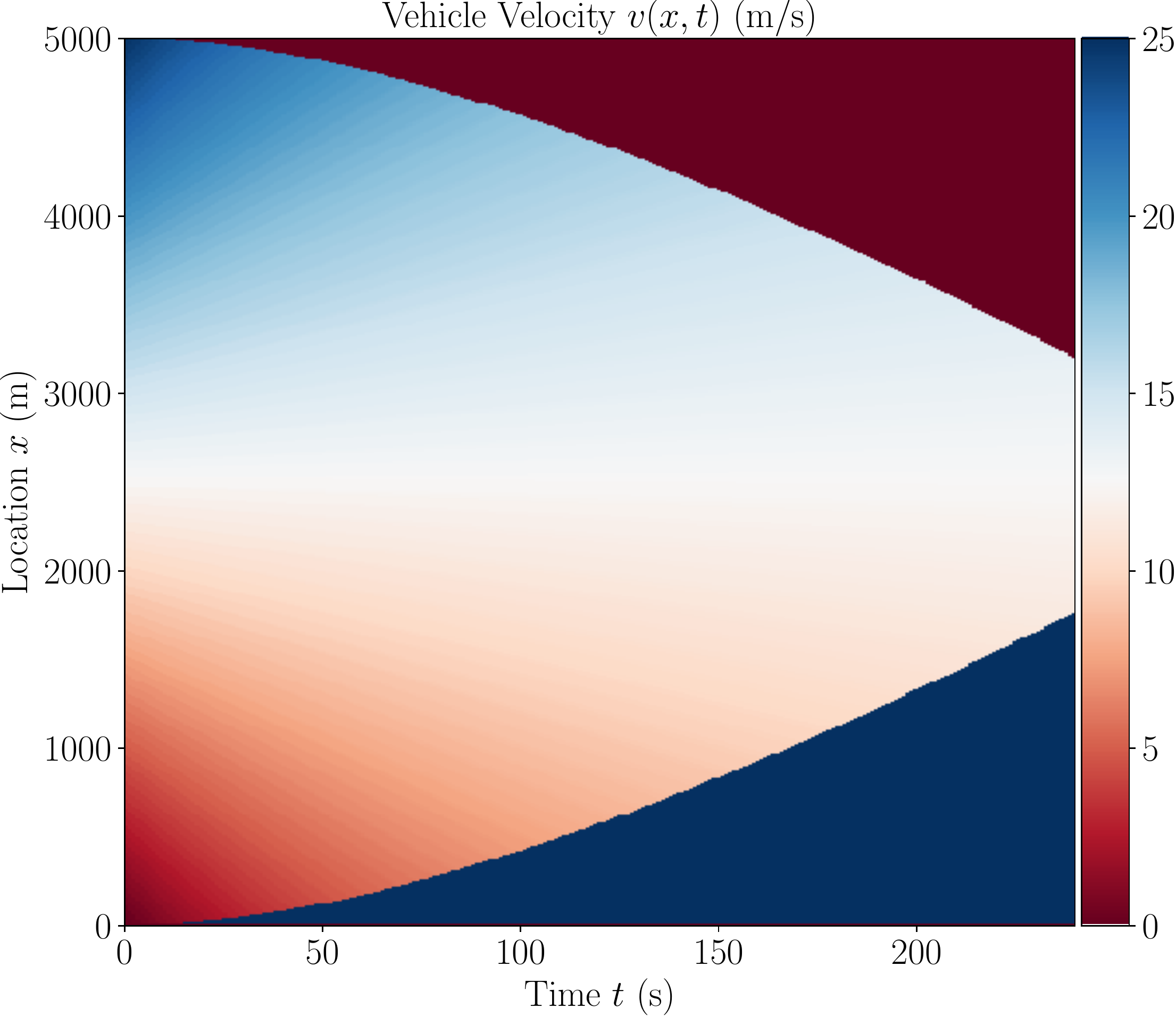}
        \caption{Dataset of Vehicle Velocity $v(x, \; t)$}
        \label{fig:data}
    \end{center}
\end{figure} 

The upstream flow at $x = 0$ and the downstream flow at $x = 5000$ are set to zero. In the synthetic dataset, the pileup of vehicular traffic is back-propagating in space and forward-propagating in time. The maximum density $\rho_{max}$ is set as 0.05 vehicles per meter, and free-flow speed $v_{free}$ is 25 meters per second. The resulting velocity dataset of $v(x, \; t)$ is shown in Fig.~\ref{fig:data}.

Both the PIDL and DL neural networks in this study are constructed with the exact learning architecture as they share identical numbers of layers and neurons. The physics of traffic flow is the unique prior knowledge that PIDL possesses.

\begin{figure}[htbp]
    \begin{center}
        \includegraphics[width=0.5\textwidth]{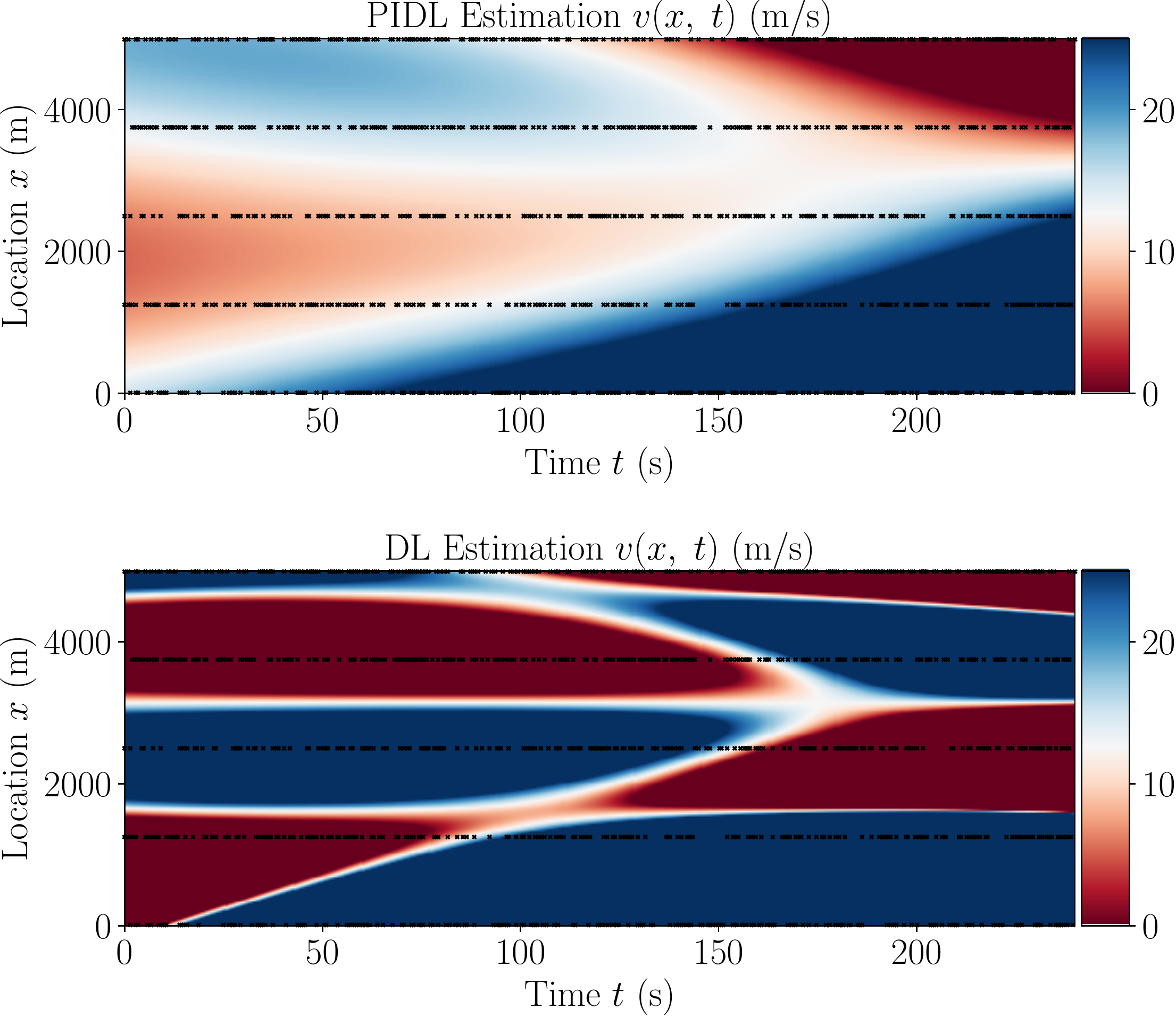}
        \caption{PIDL \& DL Estimation Result}
        \label{fig:fix}
    \end{center}
\end{figure}

We assume that loop detectors are installed at 5 locations on the 5000-meter road segment. That means the training dataset of vehicle velocity consists of sensing measurements from 5 locations throughout the 5000-meter road segment. As the temporal resolution of the dataset is 1-second, $240 \times 5 = 1200$ data points are available at these 5 locations during the 240-second case study. Accounting for the occasional data loss during signal transmission and data storage, 250 samples from the 1200 data points are randomly selected as the training dataset. Note that it is an extreme data loss scenario in which only $20.8 \%$ of the data collected is broadcast and transmitted for processing. It makes the estimation task for both the DL and PIDL neural networks significantly more challenging. Fig.~\ref{fig:fix} shows the estimation result, in which the 250 training samples are also marked.

The result produced by the PIDL neural network exhibits matching patterns shown in Fig.~\ref{fig:data} of the original dataset. With the priori knowledge of physics, PIDL neural network is capable of utilizing limited training samples for TSE. In contrast, DL neural network is not able to generate a meaningful estimation of traffic density. Table~\ref{tab:fix} details the computation time and accuracy of the two neural networks for comparison. The improved results of PIDL are highlighted in \textbf{bold}.

\begin{table}[htbp]
\centering
\caption{PIDL \& DL Performance Comparison}
\begin{tabular}{ccc}
\addlinespace[-\aboverulesep]
\cmidrule[\heavyrulewidth]{2-3}
            & \textbf{Computation Time (s)} & \textbf{Accuracy}         \\ \midrule
\textbf{PIDL}        & \textbf{8.24}         & \textbf{75.2\%}   \\ \midrule
\textbf{DL}          & 41.18                 & 24.3\%            \\ \bottomrule
\end{tabular}
\label{tab:fix}
\end{table}

Next, we examine the impact of the training sample size. Using a varying training size from 500 to 1000 (given at maximum 1200 datapoints are available at the five sensing locations), which represents $0.416 \%$ to $0.832 \%$ of data in the velocity field, the results of estimation accuracy of PIDL and DL are listed in Table~\ref{tab:fixed}. Similarly, the better outcomes are highlighted in \textbf{bold}. The results indicate that PIDL outperforms DL in terms of accuracy at all sample levels.

\begin{table}[htbp]
\centering
\caption{PIDL \& DL Performance Comparison with Varying Training Size}
\begin{tabular}{cccc}
\addlinespace[-\aboverulesep]
\cmidrule[\heavyrulewidth]{3-4}
\textbf{Sample Size} & \textbf{\% Data} & \multicolumn{2}{c}{\textbf{Accuracy}}  \\ 
& & \textbf{PIDL} & \textbf{DL}\\ \midrule
\textbf{500} &  \textbf{0.416\%} &  \textbf{81.65\%} & 31.24\% \\ \midrule
\textbf{750} & \textbf{0.625\%} &  \textbf{83.32\%} & 35.72\% \\ \midrule
\textbf{1000} & \textbf{0.832\%} &  \textbf{82.79\%} & 29.18\% \\ 
\bottomrule
\end{tabular}
\label{tab:fixed}
\end{table}

\section{Conclusion}\label{sec:C}

In this paper, we showcase PIDL for transportation applications such as for traffic state estimation. The case study shows the capability of PIDL in encoding the physical laws of traffic flow to make accurate and timely velocity estimations with limited and sparsely located measurements. The results are encouraging and pose several future research directions. For instance, how can this approach be adapted to suit the developing countries' situation with mixed traffic conditions (and a nonexplicit traffic flow law)? How can this approach be incorporated into an arterial network with signalized intersections? Future work should also consider integrating higher-order traffic flow models and demonstrate the applicability to field data. 

\bibliographystyle{unsrt}
\bibliography{application} 
\end{document}